# A survey of air combat behavior modeling using machine learning

Patrick Ribu Gorton, Andreas Strand, Karsten Brathen, *Senior Member, IEEE*

***Abstract*—With the recent advances in machine learning, creating agents that behave realistically in simulated air combat has become a growing field of interest. This survey explores the application of machine learning techniques for modeling air combat behavior, motivated by the potential to enhance simulation-based pilot training. Current simulated entities tend to lack realistic behavior, and traditional behavior modeling is labor-intensive and prone to loss of essential domain knowledge between development steps. Advancements in reinforcement learning and imitation learning algorithms have demonstrated that agents may learn complex behavior from data, which could be faster and more scalable than manual methods. Yet, making adaptive agents capable of performing tactical maneuvers and operating weapons and sensors still poses a significant challenge.**

**The survey examines applications, behavior model types, prevalent machine learning methods, and the technical and human challenges in developing adaptive and realistically behaving agents. Another challenge is the transfer of agents from learning environments to military simulation systems and the consequent demand for standardization.**

**Four primary recommendations are presented regarding increased emphasis on beyond-visual-range scenarios, multi-agent machine learning and cooperation, utilization of hierarchical behavior models, and initiatives for standardization and research collaboration. These recommendations aim to address current issues and guide the development of more comprehensive, adaptable, and realistic machine learning-based behavior models for air combat applications.**

## I. Introduction

BEHAVIOR models are fundamental components of simulation-based fighter pilot training and other air combat applications. Computer-generated forces (CGFs) are autonomous simulated entities used in military simulations to represent friendly and opposing forces [1]. Modeling their behavior to make them act realistically and human-like is demanding. Consequently, in today's simulation-based training, pilots engage in combat with CGFs that act in predictable ways and generally lack realistic behavior. Therefore, the instructors must manually control many aspects of the CGFs to ensure the pilots get the training needed [2, 3]. Freeman *et al.* [4] advocate four functional requirements to achieve realistic behavior, which are tactical inference, tactical action, modal behavior, and instructional capability.

Patrick Ribu Gorton (e-mail: patrick-ribu.gorton@ffi.no), Andreas Strand (e-mail: andreas.strand.so@gmail.com), and Karsten Brathen (e-mail: karsten.brathen@ffi.no) are with FFI (Norwegian Defence Research establishment), Instituttveien 20, 2007 Kjeller, Norway.

Besides training, realistic behavior models benefit applications like mission planning and tactics development. Modeling and simulation tools can help mission planners predict and evaluate the outcome of different scenarios, allowing refinement of strategies and tactics before the actual mission takes place [5]. Tactics development may leverage the creativity of ML agents that autonomously explore new strategies with few restrictions.

Traditionally, air combat behavior is created manually by first eliciting domain knowledge from subject matter experts (SMEs) and then creating a conceptual model reflecting this knowledge. Finally, the model is implemented in a computer program as a rule-based script, such as a decision tree or a finite state machine (FSM) [6]. This process is laborious and risks losing essential domain knowledge when transitioning from one step to another.

In recent years, there has been a surge in applying machine learning (ML) for the efficient development of air combat behavior models. Learning the behavior model with data-driven methods imposes other requirements on the model representation. The model structure should be able to capture general patterns from complex data, and at the same not be too rigid as that would restrict learning potential. A neural network is a compelling model representation because it offers scalability and parallel processing and allows iterative improvements.

The shift to ML is perhaps also inspired by advances in applying intelligent agents to complex systems in general, aided by novel approaches in reinforcement learning (RL), imitation learning (IL), and evolutionary algorithms (EAs). Notable examples in RL are DeepMind's AlphaGo and its successors, reaching superhuman levels in complex games [7, 8]. Robotic control showcases the strength of IL and how demonstrating a task is sometimes the simplest solution [9]. Furthermore, EAs excel in global optimization in high-dimensional problems [10].

The above-mentioned air combat applications employ behavior models that reflect human behavior, but this is not required in applications like unmanned combat aerial vehicles (UCAVs). Dong *et al.* [11] and Wang *et al.* [12] have surveyed maneuvering aspects of autonomous air combat, where Dong *et al.* focus particularly on lower-level control and guidance. Both surveys outlined the applications of analytical, knowledge-based, and ML methods. The examined behavior models were aimed at winning one-on-one engagements, which is a key problem in unmanned air combat and some aspects of fighter pilot training. Dong *et al.* argue that ML methods model human perceptions and decisions better, while traditional methods produce well-defined maneuvers. They also emphasize the benefit of using simulation systems to generate learning data. Wang *et al.* argue that realistic scenarios and simulations are more critical for advancing autonomous air combat than designing elaborate models, such as deep neural networks.

Løvlid *et al.* [13] have explored works related to the data-driven behavior modeling (DDBM) approach for CGFs as an alternative to the traditional approach that relies on SMEs. They highlight how the DDBM approach offers potential benefits to military end-users, such as an easier way of adding new CGF behaviors, including more human-like behaviors compared to the traditional modeling approach. However, they also point to challenges related to lack of data, noisy and incomplete data, difficulty interpreting model decisions, and striking a balance between simulation fidelity and processing performance. While their work concentrates on RL and IL in military simulation-based training and decision support systems, it encompasses a broad scope, devoting only a single reference to air combat [14].

Thus, recent contributions with air combat applications call for a revised perspective on the field. This paper presents a survey of the current state of research on behavior modeling in the air combat domain, specifically pilot behavior modeling based on ML. The purpose is to provide insight into prevalent models, methods, scenarios, and applications. Key themes emerging from the survey include the transition to multi-agent learning, alignment of simulation systems, development of benchmark scenarios, and their significance for practical applications such as pilot training. These findings shape directions for future research endeavors.

The outline of the paper is as follows. Section II describes the scope, databases, and keywords used to collect pertinent literature. Section III outlines the range of behavior model types, and Sections IV-VII introduce the prevalent ML methods used to train these. Section VIII presents the surveyed literature and key characteristics. The identified gaps and trends are discussed in Section IX followed by a conclusion in Section X with findings and recommendations.

## II. Literature Selection

This survey scopes research related to the development of realistic behavior models for air combat. The collected literature draws from several sources, including IEEE Xplore, Google Scholar, and suggestions from the Mendeley reference manager. The search terms entered on these services were *air combat*, *behavior*, *simulation*, *artificial intelligence*, and *machine learning*. Publications, reports, and theses were then subjectively filtered based on pertinence. Only literature written in English has been considered. An overview of the literature is presented in Table I.

## III. Behavior Models

A behavior model is a broad term for a model that chooses actions based on percepts and is in ML often expressed as a mathematical function. The choice of behavior representation is based on the mission task, application, previous research, and related technical requirements.

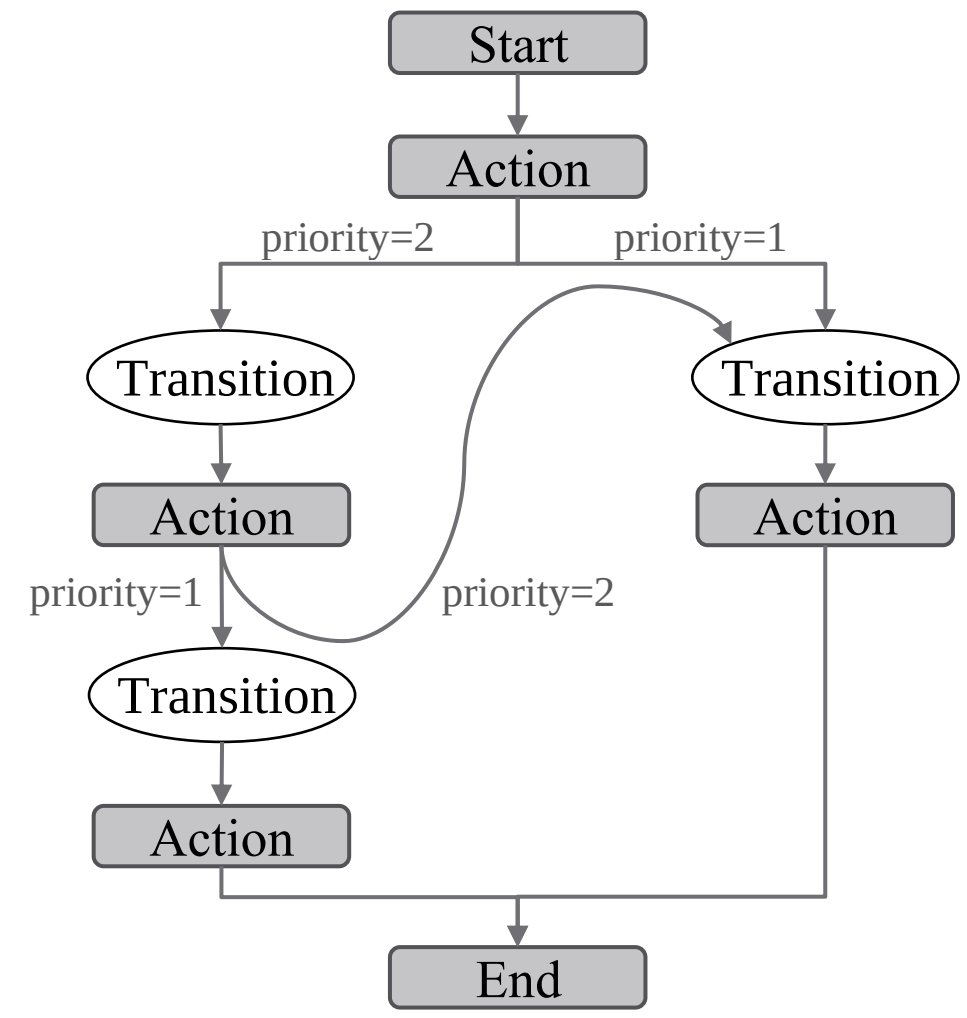

**Fig. 1.** Example of a BTN that shows how actions connected to more than one transition have assigned priorities.

### A. Behavior Transition Network (BTN)

Stottler Henke Associates developed the BTN in the early 2000s [11, 12, 13], which is essentially a behavior script represented as a graph. Actions are executed sequentially, and which action to perform next is determined by transitions.

Syntactically, BTNs are directed bipartite multigraphs. They are bipartite since all edges connect an action and a transition. A transition node expresses a condition that, when fulfilled, transitions between action nodes. Moreover, BTNs are multigraphs since action nodes may connect to multiple transition nodes expressing different conditions. In this event, the transition nodes are assigned priorities defining the order to evaluate them in case more than one condition is fulfilled. All BTNs must have a start action node and an end action node. An example of a BTN is shown in Fig. 1.

Behavior transition networks are like FSMs but include several augmentations. A BTN may be hierarchical, wherein an action node may refer to another BTN. This hierarchical property is essential to model complex behavior and avoid the state explosion of FSMs. The execution of a BTN is relatively simple, beginning at the start node and proceeding to the following node. If the following node is an action, the action gets executed. If the following node is a transition, the transition occurs only when the condition is met. If the following node branches to more than one transition, their conditions are evaluated in the order given by their priority scheme as described above. Note that in hierarchical BTNs, if an action node refers to another BTN, any subsequent transition node in the superior BTN still gets evaluated and can interrupt this sub-BTN execution. This behavior is an important aspect of BTNs that helps reduce their complexity. Finally, the execution terminates in the end action node. When controlling simulated entities, each entity will have its own BTN. The BTNs can read and write messages to and from blackboards enabling information sharing and cooperation between the entities.

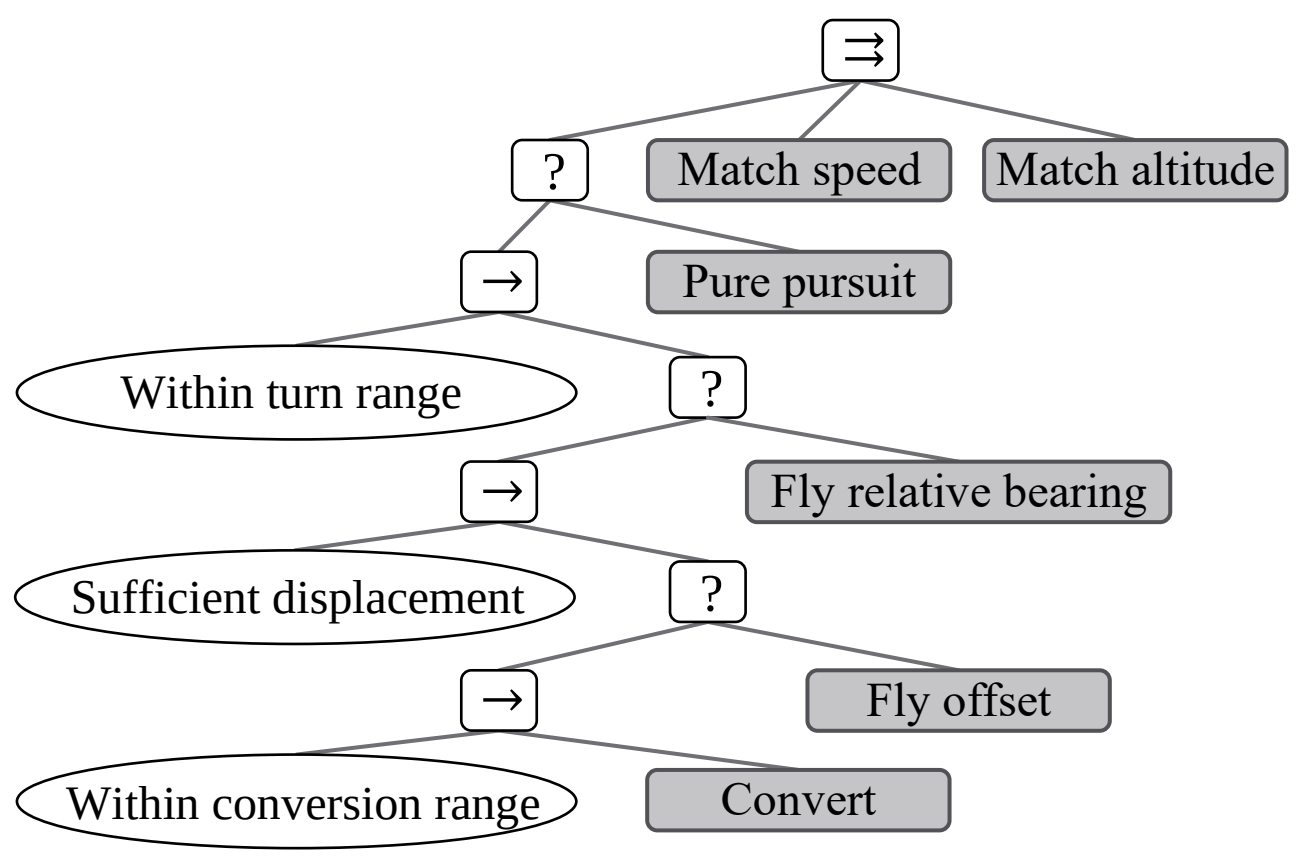

**Fig. 2.** Stern conversion intercept represented as a BT. Adapted from Ramirez *et al.* [17].

### B. Behavior Tree (BT)

Behavior trees are another paradigm for task switching that became mainstream in the 2000s for modeling non-player characters following games such as Grand Theft Auto, Halo, and Bioshock [15, 16]. A modern description of BTs [17] includes the five node types sequence (→), fallback (?), parallel (⇉), action (shaded box), and condition (oval), as demonstrated in Fig. 2. The example shows a maneuver where one aircraft intercepts another [18]. The BT gets evaluated at each tick starting at the root, which in the example is a parallel node. Consequently, the aircraft will execute three subtrees simultaneously, matching the speed and altitude of the other aircraft whilst executing the left subtree representing steering. Fallbacks execute their children from left to right until one child succeeds, while sequences execute their children until one child fails. By this logic, the aircraft will have to reach a series of sub-goals before it can finally convert.

Even though BTs span the same range of behavior models as FSMs, they are in some respects easier to manage and modify as they become complex. Because all nodes return either *success*, *failure*, or *running*, the interface is fixed and thus subtrees can be inserted anywhere in the existing model. Moreover, it makes graphical editors suitable for BTs. The flow in a BT travels down to children and then back up to parents. This is a two-way transfer of control, as opposed to the one-way transfer of control exhibited by FSMs [15].

### C. Fuzzy Tree (FT)

An FT is a fuzzy system (FS) [19] organized as a hierarchical tree structure. A defining trait of FTs is that percepts and actions are represented by linguistic variables such as *close*, *threatening*, *defensive*, or *evade*. This conforms to human reasoning, which builds on qualitative descriptions rather than numbers. The most basic components of FSs are *membership functions*, defining the linguistic variables in terms of agent percepts. In air combat with two aircraft, *close* may be represented as a sigmoid membership function that maps aircraft distance to a membership value $\mu \in [0,1]$ representing the extent to which the aircraft are *close*. The linguistic variables are used to make rules that constitute the behavior model, such as “IF enemy in pursuit AND close, perform jink”.

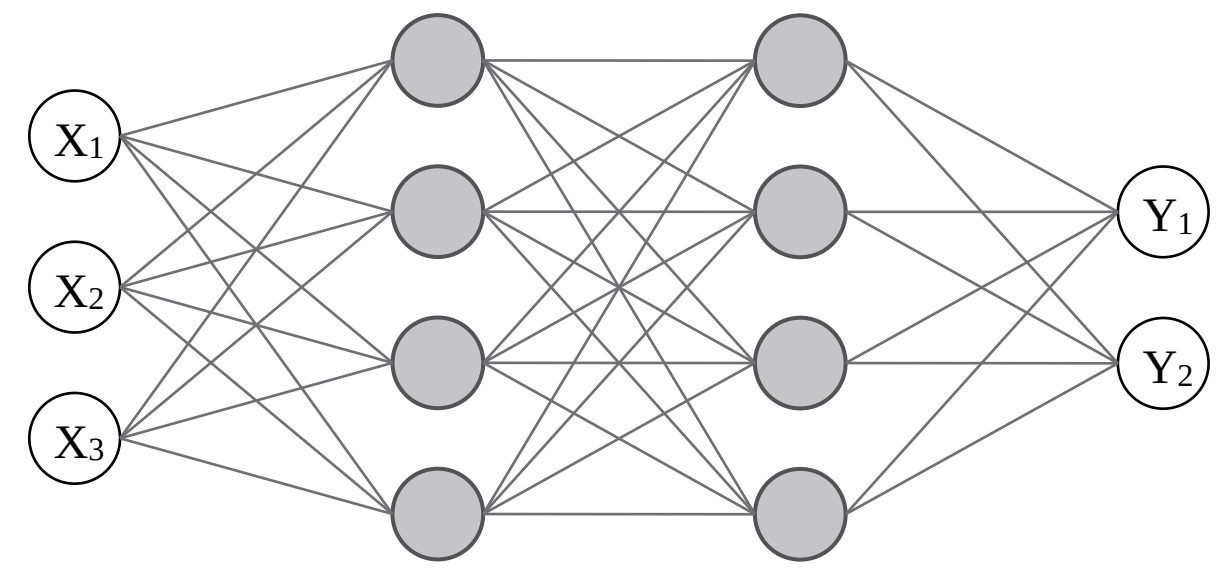

**Fig. 3.** Neural network with three percept inputs, two hidden layers of four nodes (gray), and two action outputs.

Because fuzzy logic uses numbers instead of *true* and *false*, AND returns the minimum of $\mu(\text{enemy in pursuit})$ and $\mu(\text{close})$. Whether to perform a jink depends on the membership values and a chosen threshold. If there are multiple rules for maneuvering, they must be aggregated in some way.

### D. Neural Network (NN)

Neural networks work as versatile behavior models where the rationale for the actions is implicit. The flexible structure allows a wide range of behaviors to be learned by data-driven methods. An example of an NN is shown in Fig. 3, composed of percepts $X_1$, $X_2$, and $X_3$, two hidden layers shown in gray, and actions $Y_1$ and $Y_2$. The edges in the network represent weights, and those are the trainable parameters of the behavior model. In the most basic networks, the value of each node is the weighted sum of its inputs, but modern NNs often employ some additional nonlinear transformation, which greatly increases the function space. The number and size of the hidden layers is also adjustable.

## IV. Reinforcement Learning

An agent learning by reinforcement works out what to do without being told explicitly, and this is perhaps the most common approach to learning behavior for intelligent agents. The agent interacts with an *environment* and receives *rewards* according to its performance. The aim is to maximize rewards over time by choosing the right actions [20], which amounts to finding the *optimal policy* $\pi^*$ that results in the greatest cumulative reward across all possible states.

Problems solved using RL are commonly formulated as *Markov decision processes* (MDPs), where actions influence the rewards, following environment states, and, thus, future rewards. An MDP is defined by a tuple $(S, A, T, R, \gamma)$ where

- $S$ is a set of states
- $A$ is a set of actions
- $T$ is a *transition function* specifying, for each state, action, and next state, the probability of that next state occurring
- $R$ is a *reward function* specifying the immediate reward given a state and an action
- $\gamma$ is a *discount factor* specifying the relative importance of immediate rewards

For all time steps in an MDP, the agent observes a state $s \in S$, makes an action $a \in A$ based on the state, transitions to the next state $s'$, and receives a reward $r = R(s,a)$ based on this transition. The transition function may be deterministic or stochastic, in which case $T(s,a,s') \sim P(s'|s,a)$ is the probability of the new state being $s'$ when applying action $a$ in state s. The goal of MDPs is to find the policy $\pi$ that maximizes the expected sum of discounted rewards over time

$$\arg\max_{\pi} \mathbb{E}\left[\sum_t R(s_t,a_t)|\pi\right].$$

Rewards are sometimes sparse, making it difficult for agents to learn the appropriate behavior. *Reward shaping* is a technique where supplemental rewards are provided to make a problem easier to learn, especially during the early stages of learning [21]. Agents may have to consider multiple, possibly conflicting objectives when making decisions. For example, depending on the mission, a fighter pilot must consider aspects like resource consumption, careful use of sensors to avoid being detected by opponents, and risk and safety in general. *Multi-objective* RL (MORL) is a paradigm where agents learn to balance between such priorities. In MORL, the problem usually takes the form of a multi-objective MDP, in which the reward function $R$ describes a vector of rewards, one for each objective [22].

While the above description has focused on learning the behavior of a single agent, *multi-agent* RL (MARL) studies the behavior of multiple learning agents that coexist in a shared environment. *Centralized training and decentralized execution* (CTDE) is a popular MARL framework, where agents are trained offline with centralized information and execute online in a decentralized manner [23].

There are two distinct methodological approaches to RL: searching in policy space and searching in value function space. Policy-based methods maintain explicit representations of policies and modify them through search operators. In contrast, value-based methods do not maintain policy representations but attempt to learn a value function $V^{\pi^*}(s)$, defined as the expected discounted cumulative reward for $\pi^*$ [24]. In value-based methods, the policies are implicit and can be derived directly from the value function by picking the actions yielding the best values. A third approach, actor-critic methods, approximate both policy and value functions. The *actor* refers to the policy, while the *critic* refers to the value function [20, p. 321]. Central to all RL methods is finding a balance between *exploration* of new strategies, and *exploitation*, meaning making the best decision given current information. The RL methods relevant to the survey are briefly presented here, with references to further reading.

### A. Value-based Methods

*Dynamic programming* (DP), created by Richard Bellman in the 1950s, is a class of methods for solving optimal control problems of dynamical systems and can be used to find the optimal value functions in RL problems by solving the *Bellman optimality equation* [20, p. 14]

$$\mathrm{V}^*(\mathrm{s}) = \max_{\mathrm{a}} \sum_{s'} T(s,a,s')[R(s,a) + \gamma V^*(s')]\ . \qquad (1)$$

However, classical DP algorithms are limited to trivial MDPs because of great computational expense as the number of state variables grows. As seen in (1), they assume perfect knowledge of the reward and transition functions. To circumvent these shortcomings, *approximate dynamic programming* (ADP) operates with a statistical approximation of the value function rather than computing it exactly.

The Bellman equation, with some rewriting, may also be used to find a $Q$ function, which is the value function conditioned on taking a particular action next. *Q-learning* [25] is a popular algorithm for doing this. In Q-learning, the $Q$ function is a table holding the values of all actions at all possible states and is updated by

$$Q(s_t,a_t) \leftarrow Q(s_t,a_t) + \alpha[r_t + \gamma \max_{\mathrm{a}} Q(s_{t+1},a) - Q(s_t,a_t)],$$

where $\alpha$ is the size of the update step size and $\gamma$ the factor discounting future rewards. The Q-learning algorithm operates on a finite set of discrete states and actions, and the resulting policy amounts to selecting the action that yields the greatest value given a state.

Extensions to Q-learning have been made, like *deep Q-learning* [26], achieving eminent results in human-level control applications. Deep Q-learning combines Q-learning with neural networks forming *deep Q-networks* (DQNs) for function approximation with continuous state spaces. Deep Q-learning uses a replay buffer to learn from previously collected data multiple times to improve sample efficiency and convergence. The replay buffer holds a finite but continuously updated set of transitions $(s,a,r,s')$.

Variants of Q-learning are also applicable to MARL problems. For example, under the CTDE framework, QMIX [27] is a Q-learning algorithm for learning behaviors of cooperative agents. The method employs a neural network for estimating the joint Q-values of multiple agents.

An alternative value-based RL algorithm is Monte-Carlo tree search (MCTS), a heuristic search algorithm for estimating the optimal value function by constructing a search tree using Monte-Carlo simulations [28]. This method has been used to find the best move in the board games like Go [29]. Like with ADP and Q-learning, MCTS may combine with neural networks, which in the case of Go, proved able to beat professional human players [30].

### B. Actor-Critic Methods

The *deep deterministic policy gradient* (DDPG) algorithm also adapts the ideas behind the success of deep Q-learning to the continuous action domain [31]. The actor $\pi_\theta$, which is deterministic, and critic $Q_\phi$ are neural networks parameterized by $\theta$ and $\phi$. Like in deep Q-learning, DDPG makes use of a replay buffer with $N$ transitions $(s,a,r,s')$ collected by iterations of the policy $\pi_\theta$. With these transitions, $Q_\phi$ is updated to satisfy the Bellman equation more closely with each iteration, as in deep Q-learning. The new $Q_\phi$ is used to update policy $\pi_\theta$ by ascending with the gradient

$$\frac{1}{\mathrm{N}}\sum_{t=1}^{N} \nabla_\theta Q_\phi\big(s_t,\pi_\theta(s_t)\big)\,,$$

which is computed by backpropagating through $Q_\phi$. However, the DDPG algorithm is reported to have some stabilization challenges and to be brittle to hyperparameter settings [32]. Beyond the use of replay buffers, sample efficiency can be improved by running multiple local copies of the actor in parallel and asynchronously or synchronously updating the global model, a technique employed in the Advantage Actor-Critic (A2C) algorithm [33].

Extending DDPG, the multi-agent deep deterministic policy gradient (MADDPG) algorithm [34] is an actor-critic method for MARL, which adopts the CTDE framework. During training, the centralized critic has access to the observations and actions of all agents and guides their policy updates. After training, the agents select actions based solely on their local observations.

Another method that builds on the concepts of DDPG is the *soft actor-critic* (SAC) algorithm [35]. Unlike DDPG, SAC learns a stochastic policy, i.e. a distribution $\pi_\theta(\cdot\,|s_t)$ of actions given states. To motivate action exploration, SAC aims to maximize not only the expected return but also the *entropy* of the policy, making the objective

$$\frac{1}{\mathrm{N}}\sum_{t=1}^{N}\left[R(s_t,a_t)+cH\big(\pi_\theta(\cdot\,|s_t)\big)\right],$$

where $H\big(\pi_\theta(\cdot\,|s_t)\big)$ is the entropy weighted by the coefficient $c$. Though not described here, SAC also learns both a value function and a Q function, which has proven to stabilize the learning process.

Inconveniently, updates to a policy may sometimes be detrimental. *Proximal policy optimization* (PPO) algorithms [36] employ mechanisms to prevent large diversions from the current policy configuration when optimizing. Like SAC, PPO promotes action exploration by maximizing the entropy of the stochastic policy, but more central to the method is the clipped objective $L^{\mathrm{clip}}(\theta)$ estimating the quality of actions based on rewards. To make conservative quality estimates and keep the policy updates small, $L^{\mathrm{clip}}(\theta)$ is clipped to the interval $[1-\epsilon, 1+\epsilon]$. Putting together these terms with the objective of the critic $L^{\mathrm{VF}}(\phi)$ gives the complete PPO objective

$$\frac{1}{\mathrm{N}}\sum_{t=1}^{N}\left[L^{\mathrm{clip}}(\theta)+cH\big(\pi_\theta(\cdot\,|s_t)\big)-L^{\mathrm{VF}}(\phi)\right].$$

All the actor-critic methods discussed above are types of deep RL. *Dynamic scripting* (DS) [37] is a different but well-studied RL method in surveyed literature. Initially designed for behavior generation for non-player characters in video games, DS aims to meet computational and functional requirements related to speed, effectiveness, clarity, and variety in learning behavior. Dynamic scripting produces policies in the form of behavior scripts that contain a set of behavior rules. These behavior rules are if-then statements mapping states to actions and are contained in a rulebase. Each rule in the rulebase is assigned a weight value comparable to a $Q$ value. Creating a behavior script involves selecting $n$ rules from the rulebase according to their probability of being selected, equal to each weight value divided by the sum of all weights. When an agent interacts with its environment, a *weight adjustment function* updates the weights based on the rewards the agent receives. This way, favorable rules will likely be included in upcoming scripts. The weight adjustment function keeps the total weight in the rulebase constant, meaning that when the weights of specific rules increase, the weights of the other rules decrease. Due to the probabilistic nature of dynamic scripting, all scripts generated are likely to contain different sets of behavior rules.

## V. Imitation Learning

While RL attempts to solve tasks by maximizing expected rewards, IL aims to reproduce desired behavior from demonstrations, normally given by human operators. Thus, IL can be considered a class of methods that allow transferring skills from humans to robotic or computer systems [38, ch. 1]. For some tasks, it is easier to demonstrate how they are performed than to define reward functions. Besides, expert demonstrators may have their opinions on exactly how the agent should perform the task, like in the air combat domain, where realistic behavior may mean acting in line with doctrines, tactics, techniques, and procedures. Another significant challenge in IL is the dependency of subsequent states in a demonstration, which violates the assumptions of independent and identically distributed data when minimizing the loss. This violation may lead to poor performance in theory and practice [39].

Imitation learning can be categorized into passive collection of demonstrations and active collection [40]. In the passive collection setting, demonstrations are collected beforehand, and IL aims to find a policy that mimics them. In contrast, the active collection setting assumes an interactive expert that provides demonstrations in response to actions taken by the current policy.

*Behavior cloning* (BC) is an example of IL using a passive collection of demonstrations $D=\{(s_t,\pi^*(s_t))\mid t\in 1,\dots,N\}$. It approximates the expert policy $\pi^*$ with supervised learning, minimizing the difference between the learned policy $\pi$ and expert demonstrations to create an approximate policy

$$\hat{\pi}^* = \arg\min_{\pi}\sum_{s\in D}L\big(\pi(s),\pi^*(s)\big)$$

where $L$ is a distance loss function. Applying $\hat{\pi}^*$ in states substantially different from $D$ may lead to unexpected and unwanted actions due to the lack of training data. This is a problem because slight dissonances in $\hat{\pi}^*$ may cause new trajectories leading far away from $D$. This may happen even if $D$ is not collected in sequence. However, an extensive set of demonstrations will mitigate the risk.

An active collection of demonstrations may be leveraged using methods such as *DAgger* (dataset aggregation) [39]. DAgger poses a way to query the expert for more information when $\hat{\pi}^*$ leads to states not covered by the demonstrations, followed by retraining the policy using the aggregated demonstration set.

Rather than imitating demonstrations, *inverse reinforcement learning* aims to learn the expert's intent, implicit in these demonstrations, by recovering the underlying and unknown reward function [38]. Consider a reward function parameterized by a linear combination of features

$$R(s,a)=w^T\phi(s,a),$$

where $w\in\mathbb{R}^n$ is a weight vector and $\phi(s,a): S\times A\to\mathbb{R}^n$ is a feature map. For a given $\phi(s,a)$, the goal of inverse RL is to determine $w$ and thus $R(s, a)$. Having found $R(s, a)$, $\hat{\pi}^*$ can be approximated by regular RL. Inverse RL can be suitable because the "expert" is not always optimal. Besides, a policy optimal for the expert may not be optimal for the agent if they have different dynamics or capabilities.

## VI. Evolutionary Algorithms

Evolutionary algorithms (EAs) are population-based optimization algorithms suitable for solving a range of problems, like generating control policies in robotics [41]. Evolutionary algorithms are not classified as methods of RL or IL but rather belong to a paradigm of its own. One reason is because EAs manage a population of policies rather than just one. *Genetic algorithms* (GAs) are a common type of EA inspired by the Darwinian principles of natural selection, allowing population members to interact and reproduce [42]. Genetic algorithms evaluate all policies in a population using a *fitness* function, which in the MDP formulation is equivalent to a reward function. Then, a *selection mechanism* selects some policies for recombination, creating new policies that inherit traits of their parents. Usually, the new policies undergo some degree of *mutation* to explore the search space further. This process is called *evolution*, which typically terminates when the population converges, or some termination criterion is satisfied.

*Neuroevolution* is the process of evolving neural networks using GAs. A popular method for this is the *neuroevolution of augmenting topologies* [43] (NEAT), which evolves both network structures and their parameter values.

## VII. Composite Learning

The term *composite learning* is used here to refer to strategies that partition the learning into multiple stages or partition the learning task into smaller tasks.

### A. Transfer Learning

Transfer learning refers to the idea that learning can be sped up by leveraging previous knowledge from related tasks [44]. Rather than attempting to learn a difficult task from scratch, better-performing policies can be obtained with less data and training by generalizing across different tasks. Knowledge may also be transferred between different environments or from a simulated environment to the real world [45]. The benefits of transfer learning are measured by metrics such as *jumpstart* (the initial performance of an agent on a target task), *asymptotic performance* (the improvements made to the final learned performance via transfer), and *time to threshold* (the reduction in time needed to achieve a specified threshold).

### B. Curriculum learning

Curriculum learning can be seen as a special form of transfer learning where initial tasks are used to guide the learner to perform better on the final task [46]. As knowledge is transferred between tasks, the task sequence induces a curriculum, shown to improve policy convergence and performance on difficult tasks [47]. The curriculum can be designed by altering the parameters of the environment, changing the reward function, adding constraints to the task, and more.

### C. Hierarchical Learning

Hierarchical policies attempt to tackle complex tasks by breaking them down into smaller tasks. Although such hierarchies may have multiple levels, consider a two-level policy hierarchy in which the high-level policy $\mu: s \rightarrow g$ corresponds to selecting a low-level policy $\pi_g: s \rightarrow a$ executing subtask $g$. The hierarchical learning problem is to simultaneously learn the high-level policy $\mu$ and sub-policies $\pi_g$ [40]. If formulated as an RL problem, the goal is to maximize the rewards gained when the $\mu$ and $\pi_g$'s work together.

## VIII. Literature Overview

There are several ways to organize the surveyed literature to help identify gaps and trends and provide a map of air combat behavior modeling using ML. We have grouped publications that share authors and address similar problems. These 24 groups (studies) are listed in Table I and classified by 11 properties (columns). *Year* refers to the most recent publication of the study. The *applications* are described in Section I and express the intended use of the generated behavior models. Some studies examine *multi-agent* learning, where multiple agents operate and learn in the same environment. All the surveyed studies that concern multi-agent learning employ RL.

*Beyond-visual-range (BVR)* engagements have become standard in real air combat. If enemy aircraft are too far away, they are no longer visible to the unaided eye, and the pilot must rely on long-range detection systems. The visual range on a clear day is roughly 12 km depending on the target airframe size [48]. In contrast, *within visual range (WVR)* are engagements where the enemies are visible to the pilot's naked eye.

The behavior models are learned in the context of a *mission task* assigned to one or more agents. The mission task corresponds to the mission or basic task a pilot could encounter in an operation. The behavior model expresses maneuvers or other functions of a pilot as output values. Naturally, only a subset of the pilot functions is included, and these are referred to as *agent functions*.

A key choice in modeling is the type of *behavior model* discussed in Section III, which shapes the behavior of the agent and how it will learn. The subsequent columns list the *learning paradigm* and *learning method* applied, for which summaries are provided in Sections IV-VII.

The *simulation system* is the environment in which a CGF resides. There is no universally favored simulation system, but rather a wide range of bespoke, commercial, and government-owned systems.

The last column states the degrees of freedom of a simulated aircraft. While many simulation systems technically provide 3D environments, some studies restrict the aircraft dynamics to 2D, reducing the number of maneuver variables in the agent function.

Based on the classification presented in Table I and related literature, Sections A-D present clear trends and challenges in the research field.

TABLE I
OVERVIEW OF THE SURVEYED LITERATURE IN TERMS OF ELEVEN KEY CHARACTERISTICS

| **Study** | **Year** | **Application** | **Single/ multi** | **BVR/ WVR** | **Mission task** | **Agent function** | **Behavior model** | **Learning paradigm** | **Learning method** | **Simulation system** | **Agent DOF** |
|---|---|---|---|---|---|---|---|---|---|---|---|
| Abbott *et al*. [49], Abbott [50], Abbott *et al*. [51] | 2015 | Training | Single | BVR, WVR | A/A, A/S, formation | M, SA | DT, BTN, NN | IL | BC, classifiers | DVTE, NGTS | 3D |
| Bae *et al*. [52] | 2023 | UCAV | Single | WVR | Dogfight | M | NN | RL, curriculum | SAC | JSBSim | 3D |
| Chai *et al*. [53] | 2023 | UCAV | Single | WVR | Dogfight | M | NN | RL | PPO, self-play | Bespoke | 3D |
| Ernest *et al*. [54], Ernest *et al*. [55], Ernest *et al*. [56] | 2016 | UCAV | Single | BVR, WVR | SEAD | W | Fuzzy tree/ system | RL | GA | Bespoke, AFSIM | 2D, 3D |
| Gorton *et al*. [57] | 2023 | Training | Single | WVR | CAP, flee | M | BTN | EA | NEAT (GA) | NGTS | 2D |
| Han *et al*. [58], Piao *et al*. [59], Sun *et al*. [60] | 2022 | Tactics, UCAV | Multi | BVR, WVR | A/A | M, W | NN | RL, hierarchy | A2C, PPO | Bespoke | 3D |
| Hu *et al*. [61] | 2021 | Planning | Single | BVR | Pursue, flee | M, W | NN | RL | DQL | Bespoke | 3D |
| Johansson [62] | 2018 | Training | Single | BVR | Dogfight | M, W | BT | EA | GA | TACSI | 3D |
| Källström [63], Källström *et al*. [64], Källström *et al*. [2] | 2022 | Training | Multi | BVR, WVR | A/S, CAP, recon, jam, police, coord | M, W, comms, jammer | NN | RL, curriculum | DDPG | Bespoke | 2D |
| Kong *et al*. [65], Kong *et al*. [66] | 2022 | UCAV | Multi | WVR | Dogfight | M | NN | RL, hierarchy, curriculum | DDPG, QMIX, RSAC, self-play | JSBSim, bespoke | 2D |
| Li *et al*. [67] | 2022 | Tactics | Single | WVR | Dogfight | M | NN | RL | PPO | Bespoke | 3D |
| Ludwig and Presnell [68] | 2019 | Training | Multi | WVR | Dogfight | M | BTN | RL | DS | NGTS | 2D |
| McGrew *et al*. [69] | 2010 | UCAV | Single | WVR | Dogfight | M | NN | RL | ADP | Bespoke | 2D |
| Pope *et al*. [70], Pope *et al*. [71] | 2022 | F-16 autopilot | Single | WVR | Dogfight | M | NN | RL, hierarchy | SAC | JSBSim | 3D |
| Reinisch *et al*. [72] | 2022 | Training | Multi | BVR | A/A | M, W, CM, R, jammer | BT, NN | RL, hierarchy | n/a | Bespoke | 3D |
| Sandström [73], Sandström *et al*. [74] | 2022 | Training | Single | n/a | Maneuver | M | NN | IL, transfer | BC | VBS3 | 3D |
| Selmonaj *et al*. [75] | 2023 | Training | Multi | WVR | A/A | M, W | NN | RL, hierarchy, curriculum | PPO, self-play | Bespoke | 2D |
| Sommer *et al*. [76] | 2021 | Training, CD&E | Single | BVR | Avoid SAM | M | NN | RL, transfer | Neural MCTS | CMO | 3D |
| Strand *et al*. [77] | 2023 | Training | Single | WVR | Formation | M | NN | RL | PPO | Bespoke | 2D |
| Teng *et al*. [78] | 2012 | Training | Single | BVR | A/A | M, W, CM | NN | RL | Q-learning | STRIVE | 3D |
| Toubman [3] | 2020 | Training | Multi | BVR | A/A, CAP | M, W, R, comms | Rules | RL | DS | Bespoke | 2D |
| Yao *et al*. [79] | 2015 | Training | Single | BVR | A/A | M, W, CM, R | BT | EA | GA | Bespoke | 3D |
| Zhang *et al*. [80] | 2020 | Planning | Single | BVR | SEAD | M | NN | RL | PPO | Bespoke, AFSIM | 2D |
| Zhang *et al*. [81] | 2022 | UCAV | Multi | WVR | A/A | M, W | NN | RL | PPO, self-play | Bespoke | 3D |

The table contains the following abbreviations not previously defined:

| | |
|---|---|
| A/A | air-to-air |
| A/S | air-to-surface |
| CAP | combat air patrol |
| comms | communications |
| coord | coordination |
| CM | countermeasure |
| M | maneuver |
| n/a | not applicable |
| R | radar |
| recon | reconnaissance |
| SAM | surface-to-air missile |
| SEAD | suppression of enemy air defenses |
| SA | situational awareness |
| W | weapons |
| CD&E | concept development and experimentation |

### A. Air-to-Air Combat

The most frequent learning task is air-to-air combat and in particular dogfighting. Dogfighting is close-range WVR air combat where *basic fighter maneuvers (BFM)* are used to arrive behind enemy aircraft for a favorable engagement position. It is an art that emerged naturally in World War I [82] and follows principles such as balancing airspeed and altitude, minimizing turn rates, attacking from the direction of the sun, and avoiding overshoots. In broader terms, dogfighting is a three-dimensional geometrical problem governed by the physical limitations of the aircraft and pilots.

The Defense Advanced Research Project Agency (DARPA) Air Combat Evolution (ACE) program "seeks to increase trust in combat autonomy by using human-machine collaborative dogfighting as its challenge problem" [83]. A feasibility study for ACE called the AlphaDogfight Trials invited eight companies to make dogfighting agents that would compete in a series of knockout tournaments. The AlphaDogfight Trials culminated in 2020 when the top agent was matched with an expert human pilot and won [83]. Later, developers at ACE uploaded an agent to a modified F-16 known as the Variable In-flight Simulator Test Aircraft (VISTA) and demonstrated that the agent could control the aircraft in multiple sorties with various simulated adversaries and weapons systems [84].

Air combat BVR was first seen on large scale in the Vietnam War [48], and has gradually become the main type of air combat engagement due to more advanced weapons, sensors, and sensor fusion [72, 85]. The BVR aspect adds maneuver elements such as breaking radar locks and exhausting the energy of incoming missiles. When to use radar and fire missiles becomes critical [86].

### B. Neural Networks

Three-quarters of the studies use neural networks to represent behavior models. Deep neural networks provide a hierarchical representation of the environment. The first hidden layer constitutes new abstract concepts based on the percepts. Each consecutive layer represents more abstract concepts [87]. Thus, there is potential in deep neural networks to express more advanced concepts such as threat assessment, engagement envelopes, energy management, and cooperation. The hidden layers are often powers of two to optimize memory allocation and access. Two hidden layers of 256 nodes are used in multiple studies [53, 66, 81], but generally, a large range of shapes and sizes are employed, which is necessary to accommodate different state spaces, action spaces, and behaviors.

In most of the simulations we are concerned with the agent acts based on only the observed state of the entities at the current time. However, some studies apply *gated memory* to represent the states at earlier time steps. Hu *et al.* [61] employ long short-term memory (LSTM) in their policy network to represent the percepts of 30 state variables from the five previous time steps along with the current one. Each LSTM unit decides what information to keep and what information to discard from the previous state. Bae *et al.* [52] measured a substantial increase in performance when adding an LSTM layer, especially in partially observable environments where the agent is less informed. Kong *et al.* [66] and Selmonaj *et al.* [75] use gated recurrent units, a simpler version of LSTM, in both actor and critic networks. Zhang *et al.* [81] and Källström *et al.* [2] simply make one concatenate input layer with the four most recent percepts of the state variables.

Simulations with many entities, and potentially a variable number of entities, pose a question of how to design the input layer of the policy network. Too many percepts will be a noisy representation of the environment, yet there must be sufficient percepts to enable situational awareness. Han *et al.* [58] use graph attention networks to recognize the most important ally and enemy in the air combat scenario for all aircraft. Based on that they construct an input layer representing the ownship, primary enemy, primary ally, and the primary enemy of the primary ally. Additionally, they include information on all aircraft that have the agent aircraft as the most important friend, but to fix the input length, they add these together by variable. The solution of Kong *et al.* [66] is to embed information about all friendly and enemy aircraft in multi-head attention structures, while Selmonaj *et al.* [75] opt to include only the closest friends and enemies. Zhang *et al.* [80] also discuss the challenge of a variable number of entities and suggest either representing irrelevant entities by zeros or dividing the play area into regions with associated counts of friends and enemies.

### C. Actor-Critic Methods

Many of the studies train the neural network with actor-critic methods, represented by A2C [33], DDPG [31], PPO [36], and SAC [35], which combine advantages from policy-based and value-based methods, such as allowing high-dimensional state spaces, continuous action spaces and reduced variance in policy gradient estimates.

Sharing layers between the actor and critic networks allows learning of common abstract concepts from raw percepts and reduces parameters and computation. Li *et al.* [67] make two hidden layers shared while other studies employ one shared layer [58, 59, 75]. Sharing layers between agents is also possible. If the agents have similar tasks, they may share the same neural network and receive a common reward [66]. Multiple studies [60, 2, 75] apply CTDE which relaxes the assumption of homogeneous agents but still produces coordination between them [88].

### D. Bespoke Simulation Systems

No simulation system is applied in more than three studies, and many are custom-built. Bespoke simulation systems provide precise control over entity dynamics and allow free manipulation of scenarios and integration of agents. They also speak to the need for meticulous data collection and proper management of potentially classified data. Building a new simulation system requires resources and knowledge of the domain, simulations, and software development. It is essential to conduct sufficient validation and maintenance to ensure the system's accuracy and reliability.

To our knowledge, only JSBSim [89] is open-source. Moreover, commercial-off-the-shelf choices are TACSI [90], STRIVE [91], VBS [92], and CMO [93]. Government off-the-shelf choices are NGTS [94], DVTE [95], and AFSIM [96].

## IX. Discussion

Most of the reviewed studies are motivated by improving the simulation-based training of fighter pilots. They represent a concerted effort to produce behavior models for air combat part-tasks and missions. Regardless, there is a long way to go from demonstrating adequate behavior in such tasks to successfully integrating the agents into pilot training in a way that enhances the experience. There are both technical and human challenges to be addressed.

Instructors would need easy access to agents capable of playing a certain part in scenarios they design, either from a small selection of adaptive agents or from a larger selection of specialized agents. An agent learns adaptive behavior by exposure to different situations and tasks, which requires time and a behavior model of a size and structure capable of representing the complexity. Most of the reviewed work apply random starting positions or velocities for each new episode, and some adjust other scenario parameters such as the number of friendly and hostile aircraft. Still, it is not feasible to implement or even predict all possible conditions that may occur in pilot training.

A high level of adaptivity presumes a full selection of agent functions including maneuvering, weapons, and radar controls. Half of the studies include weapons but only a few studies regard the choice of when and to use radar and countermeasures. While functions such as radar, countermeasures, and afterburner can be modeled implicitly in many cases, some training scenarios call for a pilot's explicit use of these functions.

Instructors should have the option to adjust the level of aggression or other traits of the CGF agents according to the training objective and experience level of the training audience. The behavior representation dictates how trait parameters can be included. Källström *et al.* [2] asked 25 pilots what they see as important characteristics of agents for use in training, and distinguish between *basic training*, *procedures*, and *missions*. The pilots saw agents with deterministic behaviors as essential in basic training. Moreover, procedures also call for advanced tactical agents, while missions demand doctrinal behavior on top of that. Challenging opponents were only found desirable in procedures and missions.

Källström *et al.* [2] also asked the pilots whether communication with the agents would enhance training. While communication would be convenient, teammates often understand how to react without it. Speech as a form of communication may be challenging yet feasible because it follows predefined protocols.

The technical challenges also include the transfer of agents from their learning environments to the simulation systems used by military personnel, especially if these systems are substantially different from one another. It must be possible to extract the percepts the agents rely on, and the action formats must align. Moreover, the flight dynamics model of the ML environment must be sophisticated enough to capture the dynamics relevant to the pilot training simulation [97]. But, considering the large demand for data in deep learning, lightweight simulation systems are preferred. To facilitate the transfer of agents, Strand *et al.* [77] suggest using a distributed simulation protocol [98] to enable the interaction between simulation systems of different fidelities. They also highlight the use of standard ML interfaces like Gymnasium [99] to allow rapid changes to an agent's state and action spaces and switches between ML methods.

Vectorized RL environments and experience replay can reduce the learning time manyfold, and most of the applied methods permit either or both techniques. Vectorized environments are easy to set up and utilize multiple processing units to collect experience concurrently. Experience replay is a key component for providing stability in off-policy algorithms, such as DDPG and SAC, and can also be used to emphasize particularly important experiences such as the use of weapons [71].

Despite similar requirements, few studies employ the same simulation system. This leads to a lack of standardization and benchmarking. One large effort in employing a standard environment is the Not So Grand Challenge comprising nine companies funded by the Air Force Research Laboratory [100, 101, 102, 103]. The companies develop their agents and test these in a common testbed built as a distributed simulation with government off-the-shelf software [4]. They have made agents for a series of 1v1 and 2v2 scenarios and are gradually building a library of adversary agents for pilot training. The idea is to have a digital librarian suggest agents for a certain scenario in a way that reflects and improves pilot performance [104, 105].

Air-to-air scenarios are predominant in the reviewed studies, but less than half are BVR, even though this is gradually becoming the norm as the reach of A/A missiles and sensors increases. Possibly, dogfights are still predominant because they involve the BFM a pilot learns early in training. In some sense, it is natural that agents follow the same curriculum as pilots. The studies that utilized curriculum learning exposed agents to increasingly difficult scenarios that involved gradually reducing tactical advantage [52], adding more opponents [58, 66], and making the opponents more competitive [75]. Källström *et al*. [2] highlight that curriculum learning can combat problems that arise in reinforcement learning with sparse rewards. These studies found that gradually increasing the scenario complexity resulted in increased learning efficiency.

However, it is needless to make agents learn certain elements, such as dogfighting, if this is not their intended final employment. In fact, BFM is potentially a large detour to proficiency in BVR scenarios which are generally less acrobatic and more strategic.

The fact that dogfights are less strategic and more tangible may also be a reason they are preferred as use cases. Simple goals such as avoiding being shot or shooting down enemy aircraft are easier to reflect in reward functions that guide RL properly. Nonetheless, the goals in BVR combat are not so different. Some studies actively use reward shaping to improve learning convergence and include doctrine or domain knowledge [59, 60, 3].

Multi-agent RL has emerged as a powerful learning paradigm due to its ability to capture interactions and tactical dependencies and dynamics between agents. Pilots do not operate alone, and it is essential they learn how to cooperate with their flight and squadron. In MARL, each agent represents a non-stationarity for the other friendly and adversarial agents, which makes the learning fundamentally more difficult, but also more realistic [106]. Key aspects of cooperation include formations, target coordination, and defensive support, which all become second nature to pilots eventually. In contrast, to explicitly formulate collaborative behavior rules is hard [58].

Complex behavior models may exploit hierarchical structures to break down a task into smaller parts, as described in Section VII-C. The high-level policies in the surveyed literature all include the choice between at least one defensive and one aggressive sub-policies. A defensive policy is used when the opponent has the advantage, and vice versa. Kong *et al.* [66] and Pope *et al.* [71] include a dedicated sub-policy for attaining the control zone position behind the opponent aircraft. Selmonaj *et al.* [75] and Kong *et al.* [66] also include target selection in the high-level policy. The architecture of Sun *et al.* [60] is distinct because the high-level policy has as many as 14 outputs that encode BFM macro actions. They use a low-level policy to decide the normal load factor and velocity command to apply to the selected macro action.

## X. Conclusion

The most prominent applications for air combat behavior models based on machine learning are enhancing simulation-based pilot training, mission planning, developing new tactics and strategies, and optimizing unmanned aerial combat vehicles. The reviewed studies exhibit a concerted effort to model behavior for specific air combat tasks, particularly in the simulation-based training of fighter pilots. However, despite notable progress, challenges persist in seamlessly integrating these models into comprehensive pilot training programs, presenting both technical and human obstacles that demand attention.

The desire for adaptable agents with a wide range of functions, including maneuvering, weapons, radar controls, and countermeasures, emphasizes the need for a comprehensive approach to machine learning-based behavior modeling. Certain studies also point to the importance of making more of the agent functions explicit to attain sufficient realism. The technical challenges of transferring agents from their learning environments to pilot training simulation systems underscore the importance of aligning percepts and action formats, as well as maintaining dynamics models that balance sophistication with execution speed. Standardization remains a challenge, with few studies employing the same simulation system, highlighting the need for initiatives like the Not So Grand Challenge to establish common testbeds.

After surveying the current state of the research field, we have reached four recommendations intended to aid advancements toward more comprehensive, adaptable, and realistic machine learning-based behavior models for air combat.

### *A. Emphasis on Beyond Visual Range Scenarios.*

Although dogfighting machine learning agents are impressive, they are not highly relevant in the current state of air combat. Based on our review, a reasonable shift of focus would be from WVR missions to prioritizing the development of behavior models that incorporate the complexities and strategic aspects required in BVR missions. This applies particularly to the applications of simulation-based air combat training, mission planning, and the development of new tactics and strategies.

### *B. Enhanced Focus on Multi-Agent Machine Learning and Cooperation*

Fighter pilots do not operate alone, yet the research field represents a preponderance of studies focusing on the behavior of a single agent. The effectiveness of multi-agent methods in capturing tactical dependencies and interactions among agents suggests a need for increased research in this area. Future studies are urged to delve deeper into cooperative behaviors among agents, emphasizing formations, target coordination, and defensive support. Understanding and simulating the complexities of teamwork and collaboration in air combat scenarios will contribute to more realistic air combat experiences.

### *C. Utilization of Hierarchical Behavior Models*

Adopting hierarchical structures to break down complex air combat scenarios into smaller, more manageable sub-problems is a promising direction for future research. Furthermore, a hierarchical decision-making process allows for a coherent representation of an otherwise convoluted policy. High-level policies guiding defensive and aggressive sub-policies, as observed in current literature, can be expanded and refined to address a broader range of mission tasks, ultimately enhancing the versatility and adaptability of behavior models applied to all mentioned applications.

### *D. Standardization and Collaboration Initiatives*

Considering the current lack of standardization and benchmarking in simulation systems, researchers would benefit from active participation in collaborative initiatives like the Not So Grand Challenge. Establishing common testbeds and standardized environments and scenarios facilitates cross-comparison of different behavior models and ensures that advancements in one research project can be applied and tested in others. This collaborative approach accelerates progress and contributes to developing more robust and universally applicable machine learning-based behavior models for air combat applications.

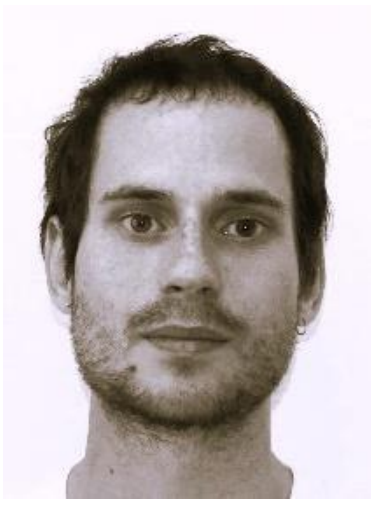

**Patrick R. Gorton** received a bachelor's degree in electronics engineering from Oslo Metropolitan University, 2018, and a master's degree in informatics specializing in robotics and intelligent systems from the University of Oslo, Oslo, Norway, 2020. He is a scientist at FFI (Norwegian Defence Research Establishment), Kjeller, Norway. His current research interests include artificial intelligence, machine learning, digital twins, and modeling and simulation, with a primary focus on behavior modeling of intelligent agents for military training and decision support.

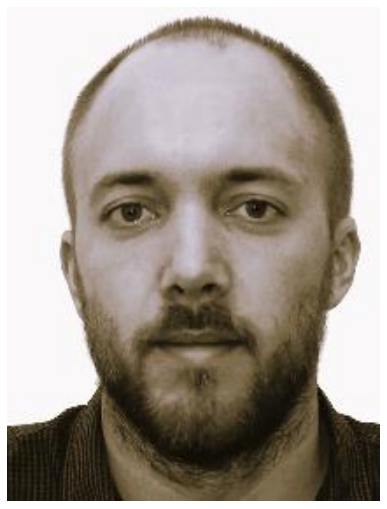

**Andreas Strand** holds a master's degree in applied physics and mathematics from the Norwegian University of Science and Technology, Trondheim, Norway, 2017, including studies in statistics at the University of California, Berkeley. He achieved a PhD in statistics from the Norwegian University of Science and Technology in 2021 on the topic of uncertainty quantification in simulations. Dr. Strand has since been engaged as a scientist at FFI, Kjeller, Norway, dedicated to behavior models for combat simulations and decision support for military operations.

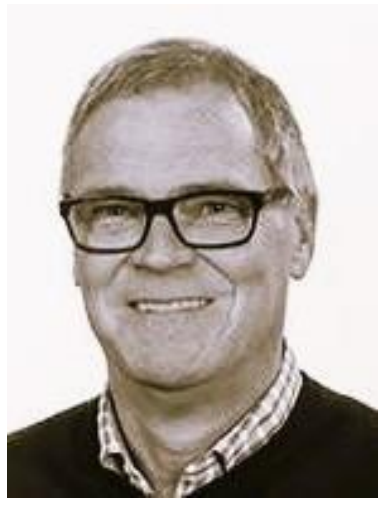

**Karsten Brathen** (Senior Member, IEEE) holds a "sivilingeniør" degree in Engineering Cybernetics from the Norwegian University of Science and Technology, Trondheim, Norway, 1979.
He has more than 40 years of experience in defense research and technology at FFI, Kjeller, Norway, and has been principal investigator and project manager for projects within submarine combat systems, high speed marine craft cockpit, naval command and control systems and defense simulation technologies. He was a visiting scientist at Ascent Logic Corporation, San Jose, CA, USA and has been an adjunct lecturer in human-machine systems engineering at the University Graduate Center, Kjeller, Norway. For many years he was the Norwegian principal member to the Modeling and Simulation Group in NATO Science and Technology Organization and the Norwegian national coordinator to the Simulation Technology capability area in the European Defense Agency. His research interests include defense modeling and simulation, behavior modeling, command and control systems and human-machine systems. Mr. Karsten Brathen is a member of ACM.